\documentclass{article}
\usepackage[accepted]{icml2020}
\usepackage[utf8]{inputenc} % allow utf-8 input
\usepackage[T1]{fontenc}    % use 8-bit T1 fonts
\usepackage{hyperref}       % hyperlinks
\usepackage{url}            % simple URL typesetting
\usepackage{booktabs}       % professional-quality tables
\usepackage{amsfonts}       % blackboard math symbols
\usepackage{nicefrac}       % compact symbols for 1/2, etc.
\usepackage{microtype}      % microtypography
\usepackage{algpseudocode}
\usepackage{algorithm,amssymb,amsthm}
\usepackage{xcolor}
\usepackage{graphicx}
\usepackage{amsmath}
\usepackage{multicol}
\algdef{SE}[DOWHILE]{Do}{doWhile}{\algorithmicdo}[1]{\algorithmicwhile\ #1} %

\newcommand{\eat}[1]{}
 {
      \theoremstyle{plain}
      
  }
\numberwithin{equation}{section}

\DeclareMathOperator*{\argmax}{argmax}

\def \Ck {{\mathcal{C}_k}}
\def \Lk {{\mathcal{L}_k}}
\def \Rk {{\mathcal{R}_k}}
\def \C {{\mathcal{C}}}
\def \L {{\mathcal{L}}}
\def \R {{\mathcal{R}}}
\newcommand*{\Comb}[2]{{}^{#1}C_{#2}}%
\icmltitlerunning{Optimal box size selection}
\begin{document}

\twocolumn[
\icmltitle{A decision-tree framework to select optimal box-sizes for product shipments}

% You can specify symbols, otherwise they are numbered in order.
% Ideally, you should not use this facility. Affiliations will be numbered
% in order of appearance and this is the preferred way.
\icmlsetsymbol{equal}{*}

\begin{icmlauthorlist}
\icmlauthor{Karthik S. Gurumoorthy}{amazon}
\icmlauthor{Abhiraj Hinge}{amazon}
\end{icmlauthorlist}

\icmlaffiliation{amazon}{Amazon, Bangalore, India}
\icmlcorrespondingauthor{Karthik Gurumoorthy}{gurumoor@amazon.com}
\icmlcorrespondingauthor{Abhiraj Hinge}{snehinge@amazon.com}

% You may provide any keywords that you
% find helpful for describing your paper; these are used to populate
% the "keywords" metadata in the PDF but will not be shown in the document
\icmlkeywords{Box size selection, Clustering, Decision-trees, Product shipments}

\vskip 0.3in

]
\printAffiliationsAndNotice{}
\begin{abstract}
In package-handling facilities, boxes of varying sizes are used to ship products. Improperly sized boxes with box dimensions much larger than the product dimensions create wastage and unduly increase the shipping costs. Since it is infeasible to make unique, tailor-made boxes for each of the $N$ products, the fundamental question that confronts e-commerce companies is: \emph{``How many $K << N$ cuboidal boxes need to manufactured and what should be their dimensions?''} In this paper, we propose a solution for the \emph{single-count} shipment containing one product per box in two steps: (i) reduce it to a clustering problem in the $3$ dimensional space of length, width and height where each cluster corresponds to the group of products that will be shipped in a particular size variant, and (ii) present an efficient forward-backward decision tree based clustering method with low computational complexity on $N$ and $K$ to obtain these $K$ clusters and corresponding box dimensions. Our algorithm has multiple constituent parts, each specifically designed to achieve a high-quality clustering solution. As our method generates clusters in an incremental fashion without discarding the present solution, adding or deleting a size variant is as simple as stopping the backward pass early or executing it for one more iteration. We tested the efficacy of our approach by simulating actual single-count shipments that were transported during a month by Amazon using the proposed box dimensions. Even by just modifying the existing box dimensions and not adding a new size variant, we achieved a reduction of $4.4\%$ in the shipment volume, contributing to the decrease in non-utilized, air volume space by $2.2\%$. The reduction in shipment volume and air volume improved significantly to $10.3\%$ and $6.1\%$ when we introduced $4$ additional boxes.
\end{abstract}
\section{Introduction}

\begin{figure}
\begin{center}
\includegraphics[width= 0.9\linewidth, height =5cm]{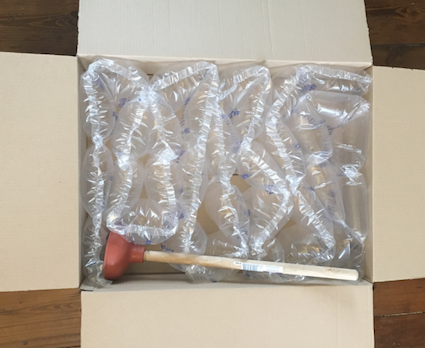}
\end{center}
\caption{Product shipped in a huge box causing excessive wastage.}
\label{fig:hugeboxsizes}
\end{figure}

E-commerce companies like Amazon often deliver their product in brown corrugated boxes. Though there is a constant strive towards package free shipping due to environmental concerns, many product characteristics like its fragility, hazardous nature, sensitivity to public disclosure (e.g. adult diapers) precludes them from being shipped without any packaging to avoid degraded customer delivery experience. In these circumstances, the best approach is to keep packaging wastage to a minimum. One of the principal contributors to such packaging wastage is the size of the packaging material, a.k.a, the box dimensions in which the products are shipped. For instance, if the box dimensions are much bigger than the product dimensions, the non-utilized empty space is often stuffed with filler material like dunnages to keep the product in position, creating added waste. The image in Fig.~\ref{fig:hugeboxsizes} drives this point home. Further, such empty spaces negatively impact the number of products that can be simultaneously transported, as the size of the individual boxes determine the quantity of shipments that can be loaded onto a container. Hence, the shipment cost per product is directly proportional to the volume of the box in which it is sent, which may be huge compared to the product volume. The ideal solution is to manufacture $N$ boxes, one for each of the $N$ products, with dimensions exactly matching the corresponding product dimension. However, this is practically infeasible due to the high fixed cost associated with making new box sizes alongside the operational difficulty involved in scaling the packaging process for a large number of box sizes, as they need to be placed in separate shelves, all in the vicinity of each other. Hence the problem of reducing the empty spaces within the box naturally breaks down into the following two sub-problems:
\begin{enumerate}
\item How many $K <<N$ boxes need to be manufactured, bearing in mind the fixed cost and operational scalability?
\item Given that $K$ boxes are manufactured, what should be their dimensions so that the overall shipment volume is minimized?
\end{enumerate}

\subsection{Contributions}
In this work, we propose an efficient algorithm to solve the problem of deciding the box dimensions of $K$ boxes. Once the sizes are determined and the boxes made accordingly, it makes practical sense to ship every product in a box that fits it snugly with minimum air volume, as it reduces both the wastage and the shipping cost. We show that given $K$, the problem reduces to a clustering problem of grouping products into $K$ clusters where each cluster specifies the set of products that will be shipped in the same sized box. Akin to $K-$means ~\cite{kmeansProof}, determining the globally optimal solution for the clusters in computationally intractable as the problem is $NP-$complete. To this end, we propose a novel, forward-backward decision tree based method to determine the $K$ clusters, which alongside simple heuristics like product-cluster reassignment and iterative dimension refinement as explained in Sec.~\ref{sec:methodology} is able to arrive at a very good local minimum of our objective namely, \emph{minimize the overall shipment volume across all product shipments}. The theoretical time complexity of our algorithm is analyzed and derived to be $O(K N \log N + N K^2)$ as discussed in Sec.~\ref{sec:computationalcomplexity}. The sub-quadratic growth rate of $O(N \log N)$ w.r.t. the number of products $N$ is critically important from the scalability perspective, as the number of products sent in boxes could potentially be in hundred-thousands in e-commerce companies like Amazon. This is the foremost advantage of our method when compared to techniques based on genetic algorithms~\cite{wongleung2006}, \cite{leeshihjia2014}. As the total shipment volume will steadily decrease with increasing $K$ from the fact that more size options are available to ship the product, the best $K$ is that value where the benefit from the decreased shipment volume is maximum compared to the cost of increased fixed cost and operational hindrance. As it may not be feasible to bring these benefits and the costs into a comparable scale, we propose to set $K$ as the \emph{elbow-point} where the decrease in shipment volume plateaus with increasing $K$, as traditionally followed in $K-$means clustering~\cite{elbow}. 

\section{Prior work}
One of the earliest references that studies the box-size problem in detail is~\cite{wilson1965}, where the problem is described as selecting the optimum number and sizes of boxes that minimizes the total shipment, warehousing and related costs.
A possibly large of set of boxes are initially created so that for every product, there exists at least one snugly fitting box where the difference between the box dimensions and the product dimensions are less than a chosen threshold $\tau$. These size variants are then consecutively eliminated till the desired number of boxes are reached. This largely heuristic-driven algorithm is not designed to optimize any objective function and hence the final sets of box sizes obtained are generally sub-optimal. In the recent past, genetic algorithms have been used to address the box-sizing problem. Specifically, Wong et. al in \cite{wongleung2006} introduce the use of multi-objective genetic algorithms (MOGA) to choose optimal box sizes for combined orders and demonstrate an application of their method to an actual industrial problem in~\cite{wongleung2008}. These methods are designed to choose box dimensions where multiple items can be packed into a single box. We henceforth refer to them as \emph{multis} where the orientation, the order, and the number of allowable items that can be packed into the same box influences the box dimensions. As majority of shipments in e-commerce conglomerates are \emph{singles} where each box holds only one product, in our present work we deal only with single-count shipments. Hence the approach developed in \cite{wongleung2008} is less useful in our setting. The MOGA technique with a problem definition very similar to ours is explored in~\cite{leeshihjia2014} under the ambit of genetic algorithms which we pit against our clustering based method in Sec.~\ref{sec:results}. The work in~\cite{leeshihjia2014} also present an optimal dynamic programming solution for one-dimensional variant of the box-sizing problem, which as explained in Sec.~\ref{sec:baseline} is used as the baseline. Generally, these evolutionary methods are very time-consuming and not scalable as many different sets of candidate solutions (box dimensions) must be evaluated individually to choose the most optimal box dimensions among them. The subsequent generations of possible dimensions are not instantiated from the view point of minimizing the overall shipment volume. Rather, they are created as minor modifications of the parent solution (crossover and mutation) and are explicitly evaluated which is computationally very expensive.

\eat{Inspired by Charles Darwin's theory of natural selection, a genetic algorithm is a search heuristic reflecting the process of natural evolution, where the fittest individuals (most promising candidate solutions) are selected for reproduction (modification) in order to produce offspring (solution set) of the next generation \cite{geneticalgo}. It applies complex principles mimicking genetic mutation, crossover, natural selection, random sampling etc. to drive optimization of the objective value (shipment volume). However, these evolutionary methods are very time-consuming and not scalable as many different sets of candidate solutions (box dimensions) must be evaluated individually to choose the most optimal box dimensions among them. The subsequent generations of possible dimensions are not instantiated from the view point of minimizing the overall shipment volume. Rather, they are generated as minor modifications of the parent solution (mutations) and are explicitly evaluated which is computationally very expensive.}

The problem for fixing the box dimensions is studied in different fields with different names. In the apparel industry, it is called the standardization problem and is framed as finding standard sizes for a given population, while minimizing the adaption loss due to mismatch in dimension. The work in this field~\cite{bongers1982}, \cite{tryfos1986}, \cite{vidal1994} focuses on solving the problem mainly for one-dimension using the distribution of the population and the interval bisection method ~\cite{vidal1992}, ~\cite{vidal1993} with different loss functions to find the optimal sizes. The box sizing can also be treated as a special case of the assortment or catalogue problem, where the goal is to optimally choose a subset from a large discrete set of possible sizes to stock, taking into consideration the space and inventory costs along with the demand for a particular size. The survey work in~\cite{pentico2008} presents a detailed review of the methodologies designed for the assortment problem in the last 50 years. In particular, it identifies the sizing problem as a special case and discusses the techniques proposed in~\cite{bongers1980} in this regard. The author in~\cite{pentico2008} notes that while~\cite{bongers1980} does present an extension for solving the sizing problem in two-dimensions, its success is highly dependent on the dimensions of the products being correlated, which does not necessarily hold in the e-commerce industry. \eat{Hence, we treat the genetic algorithm in~\cite{leeshihjia2014} as a baseline to evaluate our method developed here.}

\section{Reduction to clustering}
In order to reduce the shipment volume, it is logical that frequently sold products be shipped in size variants which are very close to their product dimension. Hence the selection of the best size variants depends on two factors: (a) the dimensions of the $N$ product that are shipped in the boxes, (b) the expected number of shipments per product, a.k.a the sales velocity. Recall that our goal is to determine the box dimensions $\{l^k, w^k, h^k\}$, $k \in \{ 1,2,\dots,K\}$  for the $K$ different size variants will be introduced. Denote $\{l_j, w_j, h_j\}$ as the product dimensions of the product $j$ for $j \in \{1,2,\ldots,N\}$ and let $s_j$ be its sales velocity. In most cases, the past shipment data can be leveraged to closely approximate $s_j$. Identifying the optimal size dimensions is tantamount to determining the set of products that will be shipped in each of the box size variant. Given any such partition of the products into $K$ clusters, let $\Ck$ denote cluster $k$ containing $N_k$ products. The cluster $\Ck$ represents the group of products that will be shipped in the same size variant $k$. Then, it is easy to see that the optimal dimensions for the box $k$, namely $\{l^k, w^k, h^k\}$ having the least shipment volume will equal the largest length, width and height of the products in $\Ck$, i.e. $l^{k} = \max\limits_{j \in \Ck} l_j$, $w^k = \max\limits_{j \in \Ck} w_j$, and $h^k = \max\limits_{j \in \Ck} h_j$. Identifying the best size variants reduces to a clustering problem, where the goal is to cluster $N$ products into $K$ clusters with the primary objective of reducing the total volume shipped. Let $x_{jk}$ be the binary membership variable determining whether product $j$ is shipped in the box $k$. The overall shipment volume can be mathematically expressed as:
\begin{align}
\label{eq:totalvolume}
&V(X) = \\
&\sum\limits_{j = 1}^{N} s_j \sum\limits_{k=1}^{K} x_{jk} \left( \max\limits_{j} x_{jk} l_{j} \times \max\limits_{j} x_{jk} w_{j} \times \max\limits_{j} x_{jk} h_j \right) \nonumber
\end{align}
where the $j,k^{th}$ entry of the binary matrix X is $x_{jk}$. Our aim is to minimize $V(X)$ subject to the binary constraints:
\begin{equation*}
\sum\limits_{k} x_{jk} = 1 \hspace{3pt} \forall j , \hspace{10pt}x_{jk} \in \{0,1\} \hspace{3pt} \forall j,k. 
\end{equation*}
As $V(X)$ strictly decreases with increasing $K$, we solve for different $K$ values and then set $K$ as the \emph{elbow-point} where we see diminishing returns with increasing $K$~\cite{elbow}.

\section{Solution methodology}
\label{sec:methodology}
Obtaining the global optimal solution for the clustering formulation in eq.(\ref{eq:totalvolume}) in the 3 dimensional space of length, width and height is computationally intractable because of the binary constraints on the membership variables $x_{jk}$. Instead, we propose the following decision tree based forward-backward algorithm to obtain a good local minimum. In each iteration, our method obtains $C$ clusters (size variants) in an \emph{incremental} fashion by using the $C-1$ clusters from the previous iteration as the starting point. Among the $C-1$ size variants, we then split one of the size variant into 2 to obtain $C$ different boxes. As the current $C-1$ size variants are akin to the leaf nodes in a binary tree out of which one is chosen to be split further, our algorithm closely resembles the decision tree based methods~\cite{Castin2018}. However as explained below, our method is composed of multiple constituents, each of them meticulously designed to minimize the specific objective in eq.(\ref{eq:totalvolume}). In the experimental section we highlight the utility of each of these parts. We obtain $K$ clusters when the algorithm completes.

Our algorithm primarily consists of 4 operations that are performed in a specific order to reach the final optimal box-dimensions. They are: (1) Cluster splitting, (2) Product reassignment, (3) Iterative refinement and (4) Cluster combination. Below we discuss these operations in detail. \eat{, whose pseudo-codes are provided in the Appendix.}

\subsection{Cluster splitting}
The process of selecting and segmenting a cluster $k$, denoted by $\mathcal{C}_k$, of $N_k$ products into two clusters of left and right child nodes, with the intention of minimizing the total volume shipped is called cluster splitting. 
\begin{figure}
\begin{center}
\includegraphics[width = 0.8\linewidth]{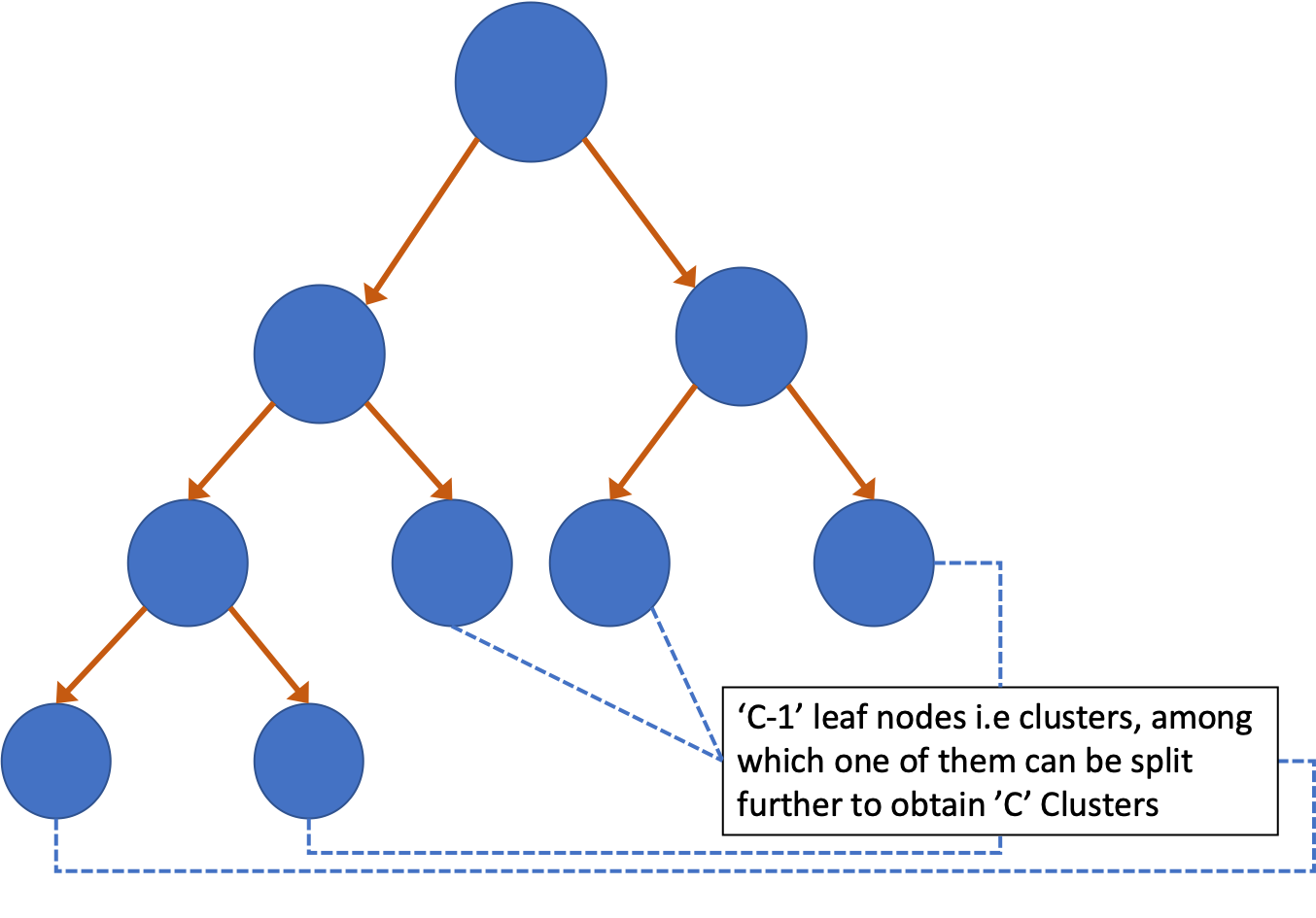}
\end{center}
\caption{Tree Created By Cluster Splitting}
\label{fig:tree}
\end{figure}
The volume for $\Ck$ is the smallest box size that can fit every product in the cluster multiplied with the number of shipments in the cluster i.e.,
\begin{align}
\label{eq:clusterVolume}
V(k) = \left( \max\limits_{j \in \Ck} l_j \right)\left(\max\limits_{j \in \Ck} w_j \right)\left( \max\limits_{j \in \Ck} h_j  \right) \left(\sum\limits_{j \in \Ck} s_j \right )
\end{align}
and the total volume across all clusters is $V = \sum_{k=1}^C V(k)$. The aim is divide \emph{one} of the current $C-1$ clusters, say $\C_k$, into two child clusters $\L_k$ and $\R_k$, assign a subset of the $N_k$ products to $\L_k$ and the remainder to $\R_k$ so that the total volume shipped is minimized. Finding the optimal solution would require examining every possible segmentation of the products, resulting in a computationally intractable time complexity of $O(2^{N_k })$. Instead, we propose a greedy search method that finds a good local minimum.

Each cluster can be split across 3 possible dimensions of length, width and height to obtain new left and right child nodes. For instance, if we decide to split the size variant $k$ containing $N_k$ products on the length $(l)$ dimension, then the objective is to decide the cut point $\tau_{k,l}$ so that the subset of products with length less than or equal to $\tau_{k,l}$ will be assigned to left node $\L_{k,l}$ and products with length greater than $\tau_{k,l}$ to $\R_{k,l}$. For each of the $C-1$ clusters and the $3$ dimensions, we will individually determine the cut points $\tau_{k,dim}$ that maximally decreases the net shipment volume. In other words, choose the cut point $\tau_{k,dim}$ that results in maximum gain, where the gain is given by: $Gain (k,dim)=V(k)-\left[V\left(L_{k,dim}\right)+ V\left(R_{k,dim}\right)\right]$ for $dim \in \{l, w, h\}$. The best dimension to split $\C_k$ is the one whose corresponding optimum cut point results in the maximum gain compared to splitting with the other two dimensions. The gain corresponding the best dimension is the gain for splitting the $\C_k$, given by: $Gain(k) = \argmax_{dim} \hspace{2pt} Gain(k,dim).$
That cluster with the maximum gain will be split into 2 on the best dimension at the optimum cut point, to obtain $C$ size variants.

The optimum cut point $\tau_{k,dim}$ for each dimension can be determined in $O(N_k \log N_k)$, by first sorting the $N_k$ products in the increasing value along that dimension and then performing a left to right sweep of $N_k-1$ possible cut point values. The usage of max-heaps for other two dimensions to keep track of the largest product dimensions in the left and right child clusters, produced for each choice of $N_k-1$ cut points, limits the complexity to be within $O(N_k \log N_k)$. Since we will be evaluating each of the $C$ clusters, the overall complexity of this step is $\sum_{k=1}^C O(N_k \log N_k) = O(N \log N)$, as $\sum_{k=1}^C N_k = N$. Our approach can be generalized to a multi-way partition of the parent cluster into more than $2$ child nodes, instead of just splitting $\Ck$ into $\L_k$ and $\R_k$. However, the time complexity will be $O(N^2)$ even for $3$-way splits and the method will not be scalable for large $N$.

\subsection{Product reassignment}
As mentioned earlier, once we have partitioned the products into $C$ clusters, the box dimensions for the cluster $\C_k$ will equal the largest length, width and height among the products in $\C_k$. However, it is possible that products are not assigned to the most optimum box that snugly fits them and minimizes the shipment volume. So for each product, we will \emph{reassign} it to that $\C_k$ whose box dimensions individually are at least as large as the product dimension and the box volume is closest to the product volume. The reassignment step is composed of iterating over each product and selecting the best  cluster $\C_k$ in terms of lowest shipment volume and involves a linear time complexity of $O(NC)$.

\subsection{Iterative refinement} 
Given a set of box dimensions, it may be possible to tweak some of them by a small amount to arrive at a new set of box dimensions that lead to more efficient packing. Iterative refinement is a process that tests out this possibility by refining the box dimensions in a greedy manner. Each iteration of the algorithm  works as follows. Assume that at iteration $t$, we have a clustering solution with the expected overall shipment volume volume $V_t$ computed as per eq.~(\ref{eq:totalvolume}). Our objective is to obtain an improved clustering solution with volume cost $V_{t+1}$ at iteration $t+1$ \emph{by moving exactly one product between two clusters} such that the difference in volume between successive iterations namely, $V_{t} - V_{t+1}$ is maximized. 
To this end, note that the dimension of any box $k$ can be changed only by moving the product with the largest length, width or height in $\C_k$ to a different cluster. So we have a maximum of three product choices per $\C_k$ and the chosen product can be moved to other $C-1$ clusters. In total, we have $3(C-1)C$ options to move one product between two clusters. We will evaluate all these $O(C^2)$ options, compute the gain in volume reduction for each of them, and greedily select the one that gives the minimal overall shipment volume $V_{t+1}$ at iteration $t+1$. If a product is moved from cluster $\C_{k_1} \rightarrow \C_{k_2}$, then then the reduction in volume equals: $V_{t} - V_{t+1} = V_t(k_1)+V_t(k_2) - \left[V_{t+1}(k_1) + V_{t+1} (k_2)\right]$, where $V_t(k)$ is the $k^{th}$ cluster volume (eq.(\ref{eq:clusterVolume})) at iteration $t$. Note that, we need to evaluate all the $O(C^2)$ combinations only for the very first iteration. For subsequent iterations, the volume reduction gains need to be computed only among $C \setminus \{k_1, k_2\} \times \{k_1,k_2\}$ and between $k_1$ and $k_2$ which are only $2C-3$ new evaluations. Each evaluation is $O(1)$, equal to the time to compute eq.(\ref{eq:clusterVolume}) with the decreased (increased) sum of sales velocity on $\C_{k_1}$ ($\C_{k_2}$) as a product $j$ with sales velocity $s_j$ is moved from $\C_{k_1} \rightarrow \C_{k_2}$, and with either the present or the second largest product dimensions in $\C_{k_1}$ depending on which dimension(s) change and perhaps new largest product dimensions in $\C_{k_2}$. The algorithm stops at iteration $T$ when all the possible moving options only increases the current shipment volume $V_t$. 

The computational complexity of this step can be computed as follows. At the beginning, we construct a max-heap for each $\C_k$ containing $N_k$ products in $O(N_k)$, one for each of the $3$ dimension, to track the products with largest dimensions which could potentially be moved to other clusters. As $\sum_{k=1}^C N_k = N$, the total pre-processing time involved is $O(N)$. Once a product is moved from $\C_{k_1} \rightarrow \C_{k_2}$, decreasing for instance the length $l^{k_1}$ of $\C_{k_1}$, we respectively delete the product from the $3$ max-heaps for $\C_{k_1}$ and push these products to the max-heaps maintained for $\C_{k_2}$, so that the products with largest dimensions in the modified clusters $\C_{k_1}$ and $\C_{k_2}$ are updated. While the delete operation in $\C_{k_1}$ for the max-heap corresponding to the length will be $O(1)$, as only the root needs to be popped out, the delete operation for other two heaps could potentially be $O(N_{k_1})$. However, the push operations into the $3$ max-heaps for $\C_{k_2}$ will all be $O\left(\log N_{k_2}\right)$. Hence the total computation complexity is $O(N+C^2 + (C + N)*T)$ where we discount $O(\log N) << N$.

\subsection{Cluster combination}
Cluster combination is the process of moving from $C$ packaging boxes to $C-1$ packaging boxes by combining the pair of clusters that produce the least additional increase in total volume shipped. We iterate over all the $\Comb{C}{2}$ possible combination and select that pair \{$\C_{k_1},\C_{k_2}$\} that gives the least total volume shipped when merged. The utility of this process may not be immediately apparent and will become clear in the next section. As we search over all possible pairs each in $O(1)$, the time complexity for evaluation is $O(C^2)$ and the final merging operation is $O(1)$.

\subsection{Final algorithm}
Having explained the constituent parts of our solution, we proceed to put these parts together and describe the actual algorithm. Recall that our objective is to find $K$ clusters that minimize eq.~( \ref{eq:totalvolume}). Our algorithm has two high-level phases, the forward pass and the backward pass. The forward pass is similar to the divisive clustering method ~\cite{ward1963}, incrementally building up the tree using clustering splitting to generate $\tilde{K} \geq K$ clusters. This process is visualized in Fig.~\ref{fig:tree}. The backward pass, following a process akin to agglomerative clustering~\cite{kaufman2009}, sequentially combines these $\tilde{K}$ clusters into the required $K$ groups. 
Creating more than the required number of clusters and then combining them in a bottom-up fashion tends to explore the solution space better leading to an improved clustering solution. For instance, let $\mathbb{C}_C$ denote the set of $C$ clusters obtained in the forward step. Say a cluster $\Ck \in \mathbb{C}_C$ is further split into $\Lk$ and $\Rk$ to get $C+1$ clusters. It could be possible to combine $\Lk$ or $\Rk$ with another cluster $\C_{\hat{k}} \in \mathbb{C}_C$ to produce a new clustering solution $\mathbb{C}^{new}_C$ of $C$ clusters which may be superior to the original solution $\mathbb{C}_C$. We test this hypothesis in Sec.~\ref{sec:experiments}, by comparing results with and without the backward pass and notice an improvement in performance in its presence. The beginning point $\tilde{K}$ for the backward pass is a hyper-parameter, chosen following the process described in Sec.~\ref{sec:tuning}. 

As the iterative refinement tries to greedily refine and improve the current clustering solution without changing the number of clusters, it is invoked following both cluster splitting and cluster combination subroutines. Whenever the box dimensions change either because of the split or merge operation, or are refined by moving products between clusters, the product reassignment step ensures that product are placed in the best-fitting box.  Thus, we perform product reassignment after each cluster split, cluster combination, and iterative refinement step. 

The forward pass starts off with one cluster, setting $C=1$ containing all the products. The dimensions of this box will equal the corresponding largest dimension among all the products. The best possible split for every cluster is evaluated using the cluster splitting method and the cluster that leads to maximum reduction in shipment volume is broken into $2$. At this point we have moved from $C$ to $C+1$ clusters. After reassigning the products to better-fitting boxes, we iterative refine and fine-tune the dimensions of the $C+1$ boxes, followed by the product reassignment step as the box dimensions may have changed. This entire procedure is repeated till we reach $\tilde{K} \geq K$ clusters. The backward pass begins at $\tilde K$ clusters where we proceed in a bottom-up fashion. After reducing the number of clusters by $1$ through merging the best two pairs using the cluster combination method, the products are reassigned, the clusters are refined by moving one product between two clusters in successive iterations to further optimize the box dimensions, followed by one more reassignment step. This agglomerative procedure is repeated till we reach exactly $K$ clusters. The maximum value of $\{l,w,h\}$ in each of the final $K$ clusters will be the dimensions of the corresponding size variants.

\subsection{Hyper-parameter selection}
\label{sec:tuning}
The only hyper-parameter in our algorithm is the beginning point $\tilde{K}$ for the backward pass. Each $\tilde{K}$ may produce different size variants once once we reach $K$ clusters from below. To choose the best $\tilde{K}$, we pursued the following validation process. We considered the actual single-count shipment data containing one product per box that occurred in a different time period, referred to as the validation set, and simulated these shipments by sending products in snugly-fit boxes whose dimensions are obtained by starting the backward pass on the training shipment set at a position $K^{\prime}$. On the simulated shipments, we then determined the percentage of air in the box $\xi$ as per eq.(\ref{eq:airinbox}) defined below. We set $\tilde{K}$ to that value of $K^{\prime}$ for which the $K$ clusters and the corresponding $K$ box sizes lead to minimum $\xi$ in the validation data set. It is important to note that the box dimensions are determined from the training set and their performance is evaluated on a different, unseen validation data set.

\section{Time complexity analysis}
\label{sec:computationalcomplexity}
Denote $\tilde{K}= \alpha K$ for some $\alpha$ independent of $N$ and $K$ and let the iterative refinement step be executed for a maximum of $T_{max}$ iterations. The time complexity for the forward pass equals:
\begin{align*}
&O\left(\sum_{C=1}^{\alpha K} N \log N + NC + N+C^2 + (C + N)*T_{max}\right) \\
&=O\left(K N \log N + N K^2 + K^3 + (K^2 + K N)*T_{max}\right).
\end{align*}
Similarly, for the backward pass it will be:
\begin{align*}
&O\left(\sum_{C=K+1}^{\alpha K} C^2 + NC + N+C^2 + (C + N)*T_{max}\right) \\
&=O\left(K^3 + N K^2 + (K^2 + K N)*T_{max}\right).
\end{align*}
As $K <<N$ and $T_{max}$ is a constant independent of $N$ and $K$, the overall time complexity can be succinctly stated as $O(K N \log N + N K^2)$.
It is worth emphasizing that the computational complexity of only $O(N \log N)$ on the number of products $N$, makes our algorithm scalable to even millions of products.
\section{Experiments}
\label{sec:experiments}
Recall that our primary goal is to decide on the number and the sizes of the boxes, so that they snugly fit the products, minimizing the non-utilized space in each shipment and thereby the overall shipment volume. In order to determine the extent of empty space ---the air in the box--- across all shipments, we use the metric $\xi$ described as follows. Let $S_{TE}$ denote the number of shipments that occurred in test time period $TE$, equal to the sum of sales velocity of the products during that interval. This interval $TE$ could be any non-overlapping period in the future, different from both the time $TR$ of the training shipments which are used to learn the box dimensions and the validation period. Given the $K$ box sizes, we first associated each product shipment $i$ with the most snugly-fitting box and computed the product and box volumes, $pv_i$ and $bv_i$ respectively. Defining $P = \sum\limits_{i=1}^{S_{TE}} pv_i$ and $V=\sum\limits_{i=1}^{S_{TE}} bv_i$ to be the total product and shipment volumes, we determine the $\%$ air-in-box, denoted by $\xi$, by:
\begin{equation}
\label{eq:airinbox}
\xi \equiv 100 \times \left(1-\frac{P}{V}\right).
\end{equation} 
As $V \geq P$, $\xi \in [0,100)$, where a value close to $0$ is indicative of the best possible box-dimensions across all products and a value near $100$ is the worst case scenario. 

As $P$ is a constant, it is clear that $\xi$ and $V$ are commensurable and minimizing $V(X)$ in eq.~(\ref{eq:totalvolume}) is tantamount is achieving smallest value for $\xi$ in eq.~(\ref{eq:airinbox}). The business sensitive nature of the shipment volumes precludes us from disclosing their actual values. Hence we report the $\%$air-in-box metric in all our experiments results. Since $\xi$ and $V$ are directly related, the inferences made using $\xi$ are straight away applicable to $V$ and vice versa.

The principal aim of our experiments is to answer the following question: \emph{``For different methods/variants, how does $\xi$ vary with $K$?''} To this end, we study the following variants of our clustering method to underscore the role played by each of different subroutines and compare it with two competing approaches.\\
(1) Our algorithm in its entirety that includes all the 4 constituent parts namely, cluster splitting, product reassignment, iterative refinement and the cluster combination involved in the backward pass.\\
(2) An alternative that comprises of the only forward pass to highlight the value addition from the backward phase.\\
(3) Exclusion of the iterative refinement step both in the forward and the backward passes.\\
(4) Another alternative that does not involve the product reassignment in both the phases.\\
(5) The Genetic Algorithm (GA) based algorithm proposed in \cite{leeshihjia2014} tailored to our setting. \\
(6) As a baseline, we also implemented the $1D$ clustering method on the product volumes as described below.
\subsection{Baseline method}
\label{sec:baseline}
Instead of clustering in the 3 dimensional space of length, width and height, we project the products into the single dimensional volumes $v_j=l_j w_j h_j$ and then cluster these $N$ volumes $\{v_1,v_2,\dots,v_N\}$ into $K$ clusters such that the following alternative objective function is minimized:
\begin{equation}
\label{eq:1D}
\tilde{V}(X) = \sum\limits_{k=1}^K \left(\max\limits_{j} x_{jk} v_j \right) \sum\limits_{j=1}^N x_{jk} s_j,
\end{equation}
subject to the binary constraints on $x_{jk}$. The one dimensional clustering formulation can be solved in $O(N^2K)$ using Dynamic Programming method \cite{clrs}, \cite{leeshihjia2014}. As before, the clustering output determines those set of products that will be shipped in a particular box variant $k$, whose dimensions will equal the largest length, width and height among the products in that cluster $\Ck$.

\subsection{Set-up and results}
\label{sec:results}
We considered about $2$ million shipments, each containing one product, that occurred during June 2019, for training. Our training data set $\mathcal{D}_{TR} = \{l_j, w_j, h_j, s_j\}_{j=1}^N$ is the set of 4-tuples for about $N=75,000$ products, containing its length $l_j$, width $w_j$, height $h_j$ and number of shipments $s_j$ known as the sales velocity. These products are currently shipped in $K=14$ boxes of different dimensions. We set July 2019 as our validation period $VD$ to determine the starting point $\tilde{K}$ as described in Sec.~\ref{sec:tuning}.

\begin{figure}
\begin{center}
\includegraphics[width = 0.95\linewidth]{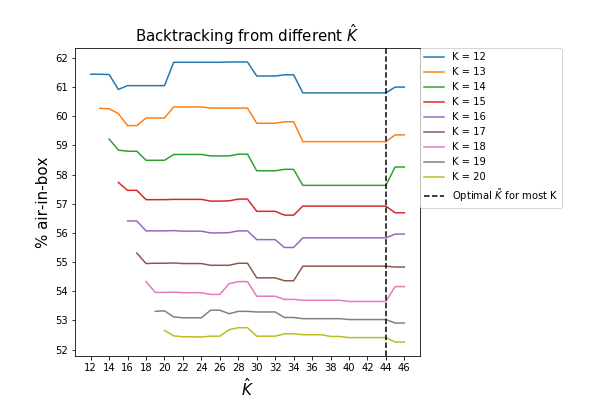}
\end{center}
\caption{The \% air-in-box on the validation set for different backtracking starting points.}
\label{fig:backtrack}
\end{figure}

We evaluated the performance of each of the $4$ different variants, the GA based approach \cite{leeshihjia2014} and the baseline method using the \%air-in-box metric $\xi$ on the test set shipments $\mathcal{D}_{TE}$,  that took place in August 2019, for values of $K \in \{12,13,\ldots, 19, 20\}$. The size of $\mathcal{D}_{TE}$ was about $2$ million shipments. For every $K$ we determined the value of $\tilde{K}$ following the process described in Sec.~\ref{sec:tuning}. The plot in Fig.~\ref{fig:backtrack} shows the $\xi$ values computed on the validation data set $\mathcal{D}_{VD}$ for different starting points $\hat{K}$, and for different $K$ box dimensions determined from the training shipments $\mathcal{D}_{TR}$. We would like to emphasize that generating the graph in Fig.~\ref{fig:backtrack} is computationally not expensive. When performing a backward pass starting from a point $\hat{K}$, the box dimensions (as a function of $\hat{K}$) for all the values of $K \leq \hat{K}$ can be obtained \emph{along the way} after merging the chosen two clusters in the group $\mathcal{C}_{K+1}$ of $K+1$ clusters to get the set $\mathcal{C}_K$ of $K$ clusters. It is \emph{not necessary} to repeat this step once for each value of $K$, corresponding to the starting point $\hat{K}$. From Fig.~\ref{fig:backtrack} we note $\tilde{K}=44$ as a good point to being the backward phase for most values of $K$.

\begin{figure*}
\centering
\begin{minipage}{.32\textwidth}
\centering
\includegraphics[width=1\textwidth]{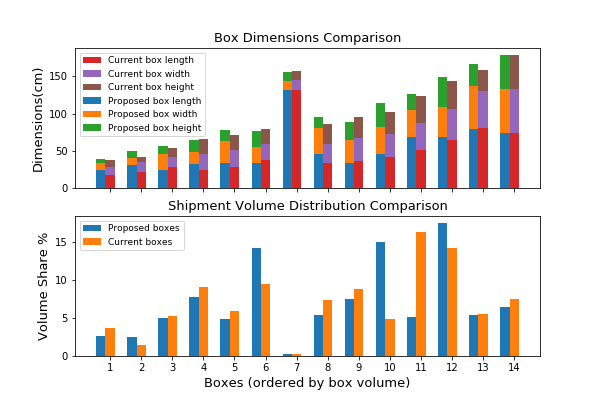}
\caption{(i) Top: Comparison of box dimensions, (ii) Bottom: Volume share distribution}
\label{fig:dimensioncomparison}
\end{minipage}
\hfill
\begin{minipage}{.32\textwidth}
\centering
\includegraphics[width= 1\textwidth]{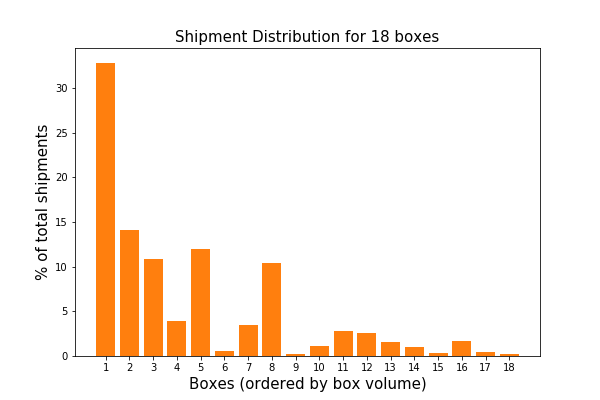}
\caption{Box usage distribution on the test shipments $\mathcal{D}_{TE}$}
\label{fig:usagedistribution}
\end{minipage}
\hfill
\begin{minipage}{.32\textwidth}
\centering
\includegraphics[width=1\textwidth]{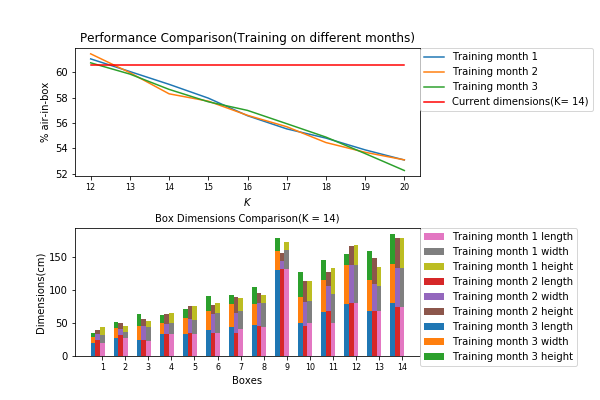}
\caption{Sensitivity of box dimensions to change in training data $\mathcal{D}_{TR}$}
\label{fig:difftrainsets}
\end{minipage}
\end{figure*}

\begin{figure}
\begin{center}
\includegraphics[width= 1\linewidth]{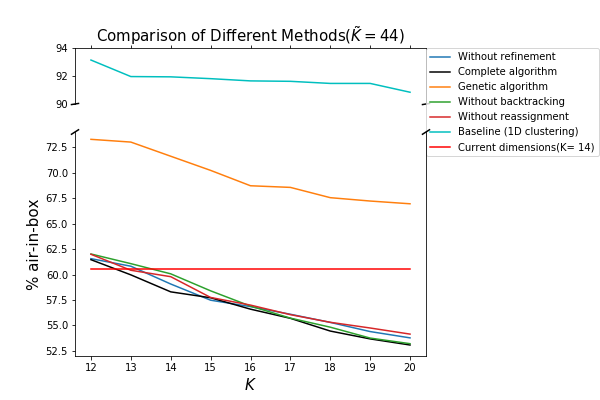}
\end{center}
\vspace{-0.2in}
\caption{Evaluation of the $4$ variants, the genetic algorithm approach and the baseline method.}
\label{fig:performance}
\end{figure}

Setting $\tilde{K}=44$, we gauged the performance of the $6$ methods. In Fig.~\ref{fig:performance} we show the \%air-in-box values computed on the test shipments $\mathcal{D}_{TE}$ for each of these methods, across different $K$ values. The horizontal line in red is the value of the metric $\xi_{curr}=60.5\%$ for the test shipments when products are shipped in the currently used $14$ box sizes. It is important to bear in mind that after manually analyzing the shipping data over several months, the current dimensions for $14$ boxes are carefully handpicked to minimize $\%$air-in-box. \emph{So any improvement over $\xi_{curr}$ is of high significance.} As expected, our complete method plotted in black, containing all the 4 sub-parts has the least $\%$air-in-box among all variants, for all values of $K$ barring $K=15$. The minor deviation at $K=15$ is because the variant without the iterative refinement step, circumstantially had a marginally better local minimum value of $57.5\%$ compared to our complete algorithm whose $\xi=57.7\%$. While each constituent part contributes to decreasing the shipment volume, the product reassignment step is the most valuable, as $\xi$ increases by more than $1\%$ in its absence. Even with $13$ boxes, \emph{$1$ less than} the current usage of $14$, our method has a lower $\xi$ value of $59.9\%$ compared to $\xi_{curr}$ and decreases further to $\xi=58.3\%$ at $K=14$. 

The baseline method, where we perform clustering in the one-dimensional projected space of product volumes, invariably performs poorly with a very high $\xi$ value of $90.8\%$ even at $K=20$ and is a poor alternative for the actual objective function in eq.(\ref{eq:totalvolume}). Though the GA based approach performs better with respect to the baseline, it consistently yields higher $\%$air-in-box values compared to our clustering based technique even after \emph{multi-starting} the method from $5$ different initial population size of $200$, where each population is a set of $K$ box sizes, and choosing the best out of the $5$ solutions based on cross-validation using the shipments in $\mathcal{D}_{VD}$. The primary reason why GA based methods may result in poor local minimum is that subsequent generations of possible dimensions are not necessarily produced from the perspective of minimizing the overall shipment volume, but are instantiated by \emph{crossing-over} the parent dimensions which could be sub-optimal.

In top half of Fig.~\ref{fig:dimensioncomparison} we ordered the boxes by their volume, and compared the dimensions of the currently used $14$ boxes against the sizes variants suggested by our algorithm. In the bottom part we show the \emph{volume share} of these boxes, where we plot $\%$volume of shipments sent in each of the size variant. By slightly increasing the box dimensions of a box $C$, our method shifts a large amount of product volume from box $C+1$, leading to smaller total shipment volume and lesser wastage of non-utilized space in the box. A prominent case of this observation is $C=10$, where by increasing the dimensions of box $10$, a huge share is taken out of the larger volume box $11$.

On simulating the actual shipments in $\mathcal{D}_{TE}$ using the $14$ box sizes produced by our method instead of the presently used size variants, we observed the overall shipment volume $V$ to decrease by $4.4\%$, translating to shipment cost savings of tens of millions of dollars even in emerging marketplaces. As the elbow point occurs at $K=18$ we recommend the usage of $18$ size variants, where we estimated the shipment volume to reduce significantly by $10.3\%$ and $\%$air-in-box by $6.1\%$ compared to the currently used $14$ box sizes. Looking into the \emph{shipment share} distribution plot in Fig.~\ref{fig:usagedistribution}, where the boxes are numbered in increasing order of their volume and we plot the $\%$ shipments sent in each of them, we notice a skewed distribution in the sense that $88.1\%$ shipments are sent in smaller boxes (number $\leq 8$) and the usage of large boxes ($\geq 9)$ are reserved only for $11.9\%$ shipments.

\subsection{Low sensitivity to training data}
Further, we analyzed the sensitivity of our algorithm to the choice of training shipments $\mathcal{D}_{TR}$, to study whether changing those leads to drastically different box dimensions. We independently executed our algorithm using $3$ non-overlapping months of shipment data as $\mathcal{D}_{TR}$ for the same hyper-parameter value of $\tilde{K}=44$, and obtained $3$ sets of $K$ box-dimensions for different values of $K$. As before, we simulated the test shipments $\mathcal{D}_{TE}$ on these $3$ sets of $K$ boxes and computed the $\%$air-in-box shown in the top-half of Fig.~\ref{fig:difftrainsets}. We observe that the $\xi$ values, across different values of $K$, vary very little over different training sets. In the bottom-half of Fig.~\ref{fig:difftrainsets}, we compare the box dimensions of the corresponding $14$ boxes obtained from each training set and again do not see any significant variations. These results strongly point to the fact that our method favorably has low sensitivity, equivalent to a \emph{low model variance} \cite{bishop2006}, w.r.t. changing the training shipments.

\section{Conclusion}
We proposed an approach for determining the sizes of the boxes used to ship products. After reducing it to a clustering problem in $3$ dimensions, we presented a decision-tree based algorithm containing forward and backward phases, coupled with steps like product reassignment, iterative refinement etc. to arrive at the best dimensions for $K$ boxes. In addition to minimizing the overall shipment volume leading to significant savings in shipment cost, our algorithm also contributes to a greener environment by keeping the wastage as low as possible. If a size variant needs to be added or deleted in the future, it is as straightforward as stopping the backward pass early or continuing it for one more iteration, \emph{as our method creates clusters in an incremental fashion without discarding the present solution.}

Extending our approach to handle \emph{multis} containing more than one product in the same shipment is a challenging task as they depend on: (i) type, the number of products and their dimensions that are shipped together (ii) the order and the orientation in which products are packed in the box. Deeper understanding of customer purchase patterns is required to identify such product groups that are bought and shipped collectively. Sparsity in the data further compounds this problem, as the number of shipments of large product groups are highly likely to be few in number. These are fruitful avenues that require further investigation.

\bibliographystyle{icml2020}
\bibliography{boxsizeoptimization}
\end{document}